\documentclass[10pt,twocolumn,letterpaper]{article}

\usepackage{cvpr}              % To produce the CAMERA-READY version
\usepackage{adjustbox}
\usepackage{array,multirow,tabularx}
\usepackage{listings,soul}
\usepackage{algorithm,algpseudocode}
\usepackage{tikz}
\usepackage{graphicx,booktabs,array,multirow}
\usepackage{xcolor,tikz}
\usepackage{colortbl}

\definecolor{SuccessTint}{HTML}{FFFFFF}
\definecolor{SuccessInk}{HTML}{252B31}
\definecolor{FailureBlue}{HTML}{6FA8DC}
\colorlet{FailureZero}{FailureBlue!10!white}
\definecolor{FailureInk}{HTML}{252B31}
\colorlet{FailureOrange}{FailureBlue}

\definecolor{AblationSeparator}{HTML}{B7C2CB}
\newcommand{\AblationConfigSep}{%
  \hspace{2.35pt}%
  {\color{AblationSeparator}\vrule width 0.3pt}%
  \hspace{2.35pt}}
\newcommand{\AblationTaskSep}{%
  \hspace{2.85pt}%
  {\color{AblationSeparator}\vrule width 0.3pt}%
  \hspace{2.85pt}}

\newlength{\AblationSRWidth}
\newlength{\AblationFWidth}
\newcolumntype{A}{>{\columncolor{white}[0pt][0.2pt]}c}
\newcommand{\AblationColumnMax}{-1}
\newcolumntype{S}[1]{>{\def\AblationColumnMax{#1}\columncolor{white}[0pt][0.2pt]}c}
\newcolumntype{B}[1]{>{\def\AblationColumnMax{#1}}c}
\newcommand{\AblationValueFont}[1]{%
  \mdseries
  \ifdim#1pt=\AblationColumnMax pt\relax\bfseries\fi}
\newcommand{\AblationCell}[3]{%
  \edef\AblationApplyColor{\noexpand\cellcolor{#1}}%
  \AblationApplyColor
  \makebox[#2][c]{#3}}

\newcommand{\SuccessCell}[1]{%
  \AblationCell{SuccessTint}{\AblationSRWidth}%
    {\color{SuccessInk}\AblationValueFont{#1}\number\numexpr10*#1\relax}}

\newcommand{\AverageCell}[1]{%
  {\color{SuccessInk}\AblationValueFont{#1}#1}}
\newcommand{\FailureNumber}[1]{%
  \ifnum#1>0\relax%
    \raisebox{0.3pt}{\fontsize{7.5pt}{7pt}\selectfont\mdseries\upshape
      \color{FailureInk}#1}%
  \fi}
\newcommand{\FailureCell}[2]{%
  \pgfmathtruncatemacro{\AblationIntensity}{10*(#2)}%
  \AblationCell{FailureBlue!\AblationIntensity!FailureZero}{\AblationFWidth}%
    {\FailureNumber{#2}}}
\newcommand{\TaskTwo}[3]{%
  \SuccessCell{#1}&\FailureCell{#1}{#2}&\FailureCell{#1}{#3}}
\newcommand{\TaskThree}[4]{%
  \SuccessCell{#1}&\FailureCell{#1}{#2}&\FailureCell{#1}{#3}&\FailureCell{#1}{#4}}

\newcommand{\HeaderTwo}{%
  \makebox[\AblationSRWidth][c]{SR$\uparrow$}&\makebox[\AblationFWidth][c]{F1}&\makebox[\AblationFWidth][c]{F2}}
\newcommand{\HeaderThree}{%
  \makebox[\AblationSRWidth][c]{SR$\uparrow$}&\makebox[\AblationFWidth][c]{F1}&\makebox[\AblationFWidth][c]{F2}&\makebox[\AblationFWidth][c]{F3}}

\usepackage{paralist}
\definecolor{cvprblue}{rgb}{0.21,0.49,0.74}
\usepackage[pagebackref,breaklinks,colorlinks,allcolors=cvprblue]{hyperref}
\def\confName{CVPR}
\def\confYear{2027}

\title{Many Senses, One World: Learning Hierarchical Multimodal World Representations for Bimanual Dexterous Manipulation}

\author{
Jin Yang, Youlin Wang, Ping Wei, \\
National Key Laboratory of Human-Machine Hybrid
Augmented Intelligence\\
Institute of Artificial Intelligence and Robotics, Xi'an Jiaotong University\\
{\tt\small \{jin.yang, pingwei\}@xjtu.edu.cn}
}

\begin{document}
\maketitle
\begin{abstract}
Bimanual dexterous manipulation relies on multimodal inputs that capture visual context and non-visual physical interaction cues for precise and coordinated control. To address the challenge of effectively integrating these modalities for policy learning, we propose HiMoDex, a two-stage framework centered on hierarchical multimodal world-state representations. In Stage 1, we pretrain a hierarchical representation network on the collected dataset of 1.19 million synchronized multimodal time steps. The network integrates visual context with tactile, force-torque, and audio cues into unified world-state embeddings. In Stage 2, we freeze the pretrained network and use its embeddings to condition a shared world-action transformer. The transformer jointly learns action generation and action-conditioned future-state prediction. Experiments across five real-world tasks show that HiMoDex achieves an average success rate of 78.7\%, representing an absolute improvement of 32.0\% over the strongest evaluated baseline. Extensive ablations validate the benefits of hierarchical representation learning and two-stage training. Failure analysis helps explain performance differences across variants. Code, data, and demos will be released at website.

% \href{https://anonymous.4open.science/w/research-website-BAB4/}{website}.

% \blfootnote{$^*$ Corresponding author} 
\end{abstract}    
\section{Introduction}
\label{sec:intro}

\begin{figure}[tbp]
  \centering
  \includegraphics[width=0.95\linewidth]{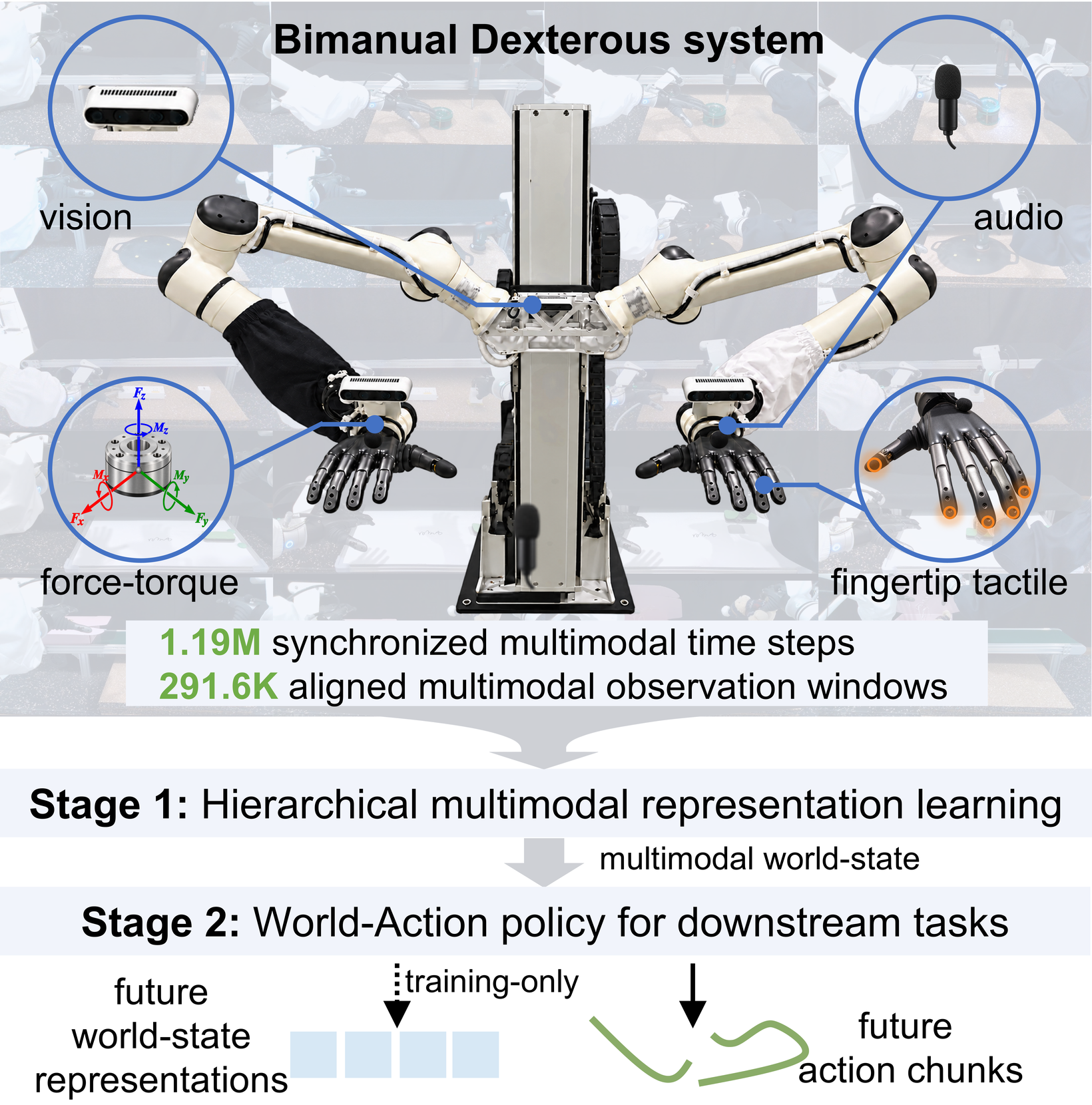}
  \caption{Our bimanual dexterous system integrates multi-view vision, fingertip tactile, wrist force-torque, and audio. We collect 1.19M synchronized multimodal time steps, organized into 291.6K aligned observation windows for multimodal world-state representation learning (Stage 1). The frozen pretrained encoders are then used for downstream policy learning (Stage 2).
  }
  \label{fig:introduction}
\end{figure}

Bimanual dexterous manipulation is essential for robots to perform real-world tasks, enabling coordinated actions, tool use, and fine manipulation~\cite{shaw2025bidex}. 
Building on advances in vision-language models and world models, vision-language-action (VLA) models~\cite{black2025pi0,physicalintelligence2025pi05,kim2025openvla} and world-action models (WAM)~\cite{yuan2026fastwam,lin2026jepawam} have made progress in robotic manipulation. 
Early methods primarily focused on parallel-jaw grippers, while recent methods have extended these models to bimanual systems with multi-fingered hands~\cite{luo2026beingh05}. 
However, these methods face challenges in contact-rich tasks requiring precise control, due to limited physical interaction information from vision alone~\cite{yu2025forcevla}.
% However, these methods still face challenges in contact-rich tasks requiring precise control, due to visual occlusion and limited physical interaction information from vision alone~\cite{yu2025forcevla}.

Tactile, force-torque, and audio feedback provide complementary cues about physical interaction. Tactile sensing captures hand-object contact, while force sensing measures interaction forces and torques. Audio can reveal impacts, friction, and events such as switch activation~\cite{liu2025maniwav}.  Previous methods have incorporated subsets of these modalities into vision-based policies to improve manipulation reliability~\cite{li2023seehearfeel,lin2025hato,yu2025forcevla}. However, jointly exploiting tactile, force-torque, and audio feedback for bimanual dexterous manipulation remains underexplored. A key challenge is to integrate heterogeneous interaction cues with visual context into a unified world-state representation for policy learning.

To this end, we build a bimanual dexterous system with multi-view vision, fingertip tactile sensing, wrist force-torque sensing, and audio. We introduce \textbf{HiMoDex}, a two-stage multimodal world-action learning framework for bimanual \textbf{Dex}terous manipulation, with \textbf{Hi}erarchical \textbf{M}ultim\textbf{o}dal world-state representation learning as its core, as shown in Fig.~\ref{fig:introduction}. In Stage 1, we learn world-state representations from aligned multimodal observation histories through hierarchical encoding and fusion. Multi-view visual histories are encoded into visual context embeddings, while tactile, force-torque, and audio histories are encoded and fused into physical interaction embeddings. A world-state encoder then integrates these intermediate embeddings into unified world-state embeddings. We train this representation network using three complementary objectives: cross-representation consistency, signal reconstruction and change prediction, and task-state contrastive learning.

In Stage 2, we freeze this representation network and use its world-state embeddings for policy learning. The policy combines these embeddings with proprioceptive histories and uses task instructions to guide action generation. A shared world-action transformer jointly learns action generation and action-conditioned future world-state prediction. The future prediction objective provides auxiliary supervision for policy learning. At deployment, only the action generation pathway is retained, avoiding the overhead of future world-state prediction~\cite{yuan2026fastwam,lin2026jepawam}.

We evaluate HiMoDex on five real-world tasks covering bimanual coordination, precise manipulation, coordinated tool use, and sustained contact. HiMoDex achieves an average success rate of 78.7\%, an absolute improvement of 32.0\% over the strongest evaluated baseline. We further conduct extensive ablations to assess the contributions of key components. Task-specific failure statistics further characterize the performance differences across variants. Our main contributions are as
follows:
\begin{compactitem}
	\item [1.] We propose a hierarchical multimodal world-state representation learning method that integrates visual context with tactile, force-torque, and audio interaction cues into unified world-state embeddings. 
    % We develop a bimanual dexterous system equipped with vision, tactile, force-torque, and audio sensors.
	\item [2.] We introduce HiMoDex, a two-stage framework using frozen pretrained encoders to learn world-action policies for bimanual dexterous manipulation.
	\item [3.] HiMoDex outperforms the strongest baseline by 32.0 percentage points in average success rate. Ablations and failure analysis clarify component contributions and performance differences.
\end{compactitem}

% To evaluate HiMoDex, we conduct real-world experiments covering a range of bimanual manipulation capabilities, including basic bimanual coordination, high-precision manipulation, coordinated tool use, and sustained-contact manipulation. HiMoDex achieves the highest average success rate among the evaluated methods, outperforming the strongest baseline by 32\%. We further conduct extensive ablations to assess the effectiveness of key components. In addition, we analyze the distribution of failure causes across ablations to identify the main sources of performance degradation. Our main contributions are as follows:

\section{Related Work}
\label{sec:relatedwork}

\begin{figure*}[tbp]
  \centering
  \includegraphics[width=0.93\linewidth]{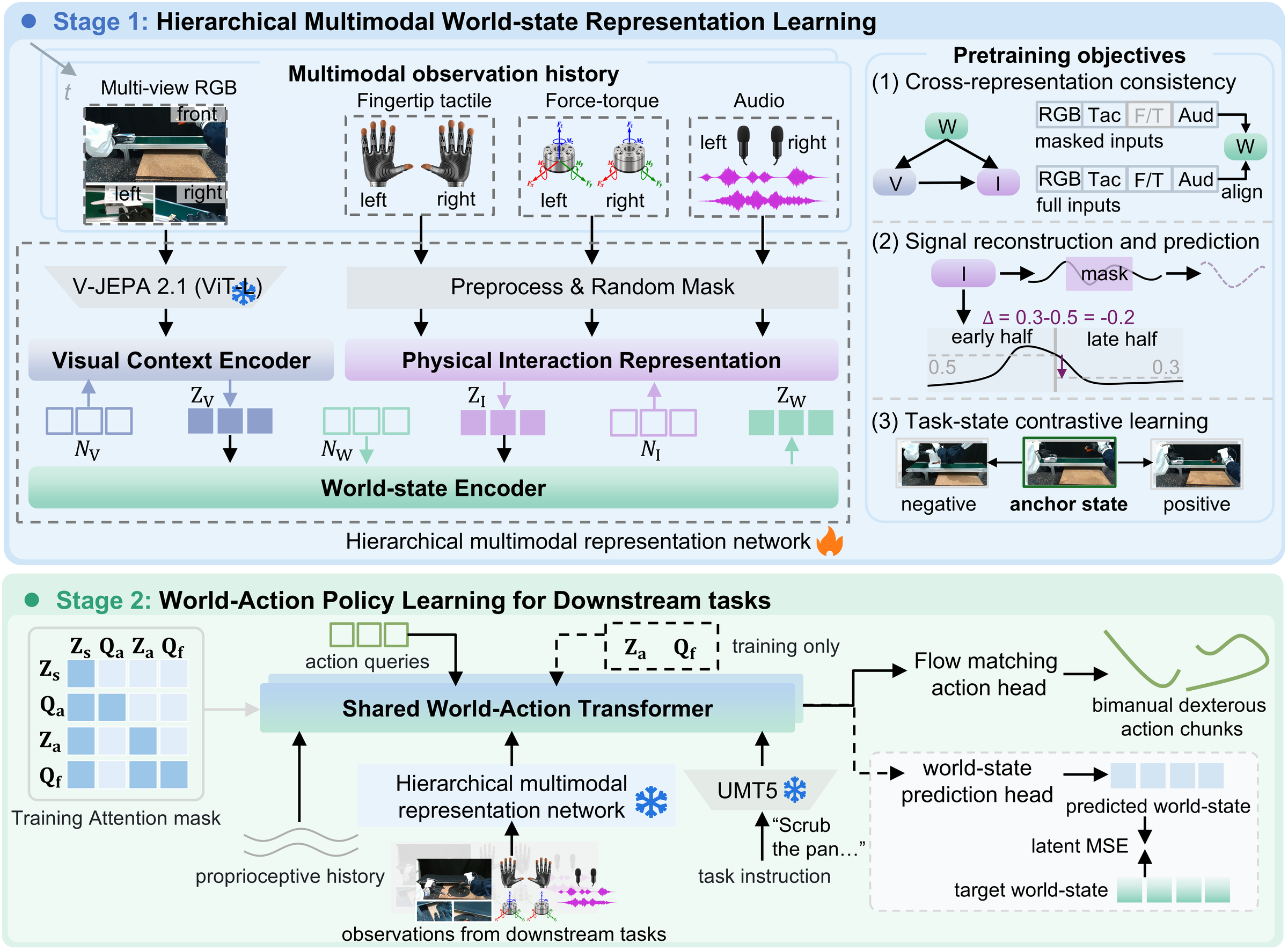}
  \caption{Overview of the two-stage training framework of HiMoDex. 
  \textbf{Stage 1} pretrains a hierarchical multimodal representation network using multi-view RGB, fingertip tactile, force-torque, and audio observation histories.
  % \textbf{Stage 1} learns the hierarchical world-state representation from multi-view RGB, fingertip tactile, force-torque, and audio observation histories.
  \textbf{Stage 2} trains a shared world-action transformer for downstream tasks, combining flow-matching action generation with future world-state prediction.
  % for flow-matching action generation and auxiliary future world-state prediction.
    }
  % Pretraining combines cross-representation consistency, signal reconstruction and prediction, and task-state contrastive learning.
  \label{fig:pipeline}
\end{figure*}

% We review the related work from three aspects.

\textbf{Multimodal Representation Learning for Robotic Manipulation.} Multimodal representation learning integrates heterogeneous observations to support robot decision-making and control.
Some studies use language supervision to learn representations that support manipulation policies~\cite{kang2025cliprt,feng2026anytouch2,ma2026fgcltp,yang2026sota}.
Specifically, FuSe~\cite{jones2025fuse} and OmniVTLA~\cite{cheng2025omnivtla} align visual and tactile observations with language descriptions to establish semantic correspondences across modalities.
Beyond language supervision, paired multimodal observations provide self-supervised learning signals.
Visuo-tactile pretraining learns cross-modal correspondences from spatially paired observations~\cite{kerr2023ssvtp}.
Masked reconstruction~\cite{krohn2026msdp} and self-distillation~\cite{higuera2025sparshx} learn robust multimodal representations, while predicting contact events or future observations~\cite{liu2025vitamin,heng2026vitacformer}. HiMoDex introduces a hierarchical representation learning approach that integrates visual context and non-visual interaction cues into a unified world-state representation.
This representation is shared by action learning and future-state prediction.

\textbf{Bimanual Dexterous Manipulation.} Compared with bimanual systems using parallel-jaw grippers, those equipped with multifingered hands enable more versatile grasping. Their higher-dimensional action spaces make coordinated control more challenging~\cite{shaw2025bidex}. 
To address this challenge, existing methods learn visuomotor policies from expert demonstrations~\cite{cheng2025opentelevision,wang2024dexcap,jiang2025dexmimicgen,luo2026beingh05}. However, visual observations provide limited information about physical interactions.
Recent studies therefore incorporate tactile sensing to complement visual observations~\cite{ding2025bunnyvisionpro,wang2026robotacdex,li2026deco,liu2026star}.
For example, HATO~\cite{lin2025hato} incorporates fingertip tactile feedback into diffusion policies to improve robustness in bimanual block stacking. VTAO-BiManip~\cite{sun2025vtaobimanip} uses visual-tactile-action pretraining for bimanual bottle-cap unscrewing. In addition to tactile feedback, MoDE-VLA~\cite{tang2026modevla} incorporates force observations into a VLA model for contact-rich bimanual dexterous manipulation.
In this work, HiMoDex fuses visual, tactile, force, and audio feedback for coordinated control across diverse dexterous manipulation tasks.

\textbf{World-Action Models for Robotic Manipulation.} World-action models extend world models by coupling future-state prediction with action generation. Based on their prediction targets, existing methods can be broadly categorized into generative approaches and latent-space approaches.
Generative approaches couple future observation prediction with action learning~\cite{wu2024gr1,guo2024pad,yuan2026dextacwam}. For example, DreamZero~\cite{ye2026dreamzero} jointly generates future videos and actions using a pretrained video diffusion backbone. 
Latent-space approaches instead predict or align future representations to avoid costly reconstruction of action-irrelevant observation details~\cite{luo2026beingh07,fang2026lewam,clemente2026lajepa}.
JEPA-WAM~\cite{lin2026jepawam} couples action learning with prediction of a joint representation of current and future observations. 
Beyond the prediction target, the deployment efficiency of WAMs also depends on whether future prediction is performed at inference. Fast-WAM~\cite{yuan2026fastwam} demonstrates that training-time future prediction improves policy performance under action-only inference~\cite{lin2026jepawam,clemente2026lajepa}. In this work, HiMoDex adopts a latent-space approach, jointly learning action generation and future world-state prediction through a shared world-action transformer.
\section{Methodology}
\label{sec:methodology}

\subsection{Problem Formulation.}

% We learn a bimanual dexterous manipulation policy $\pi_\theta$ from expert demonstrations. We model the policy as a conditional distribution $\pi_\theta(\mathrm A_t\mid l,\mathcal O_t,\mathrm P_t)$, where $l$ is a natural-language task instruction. $\mathcal O_t$ and $\mathrm P_t$ denote the multimodal observation and proprioceptive histories, respectively. Each action chunk $\mathrm A_t = [a_t,\ldots,a_{t+H-1}]$ contains $H$ consecutive actions. The observation history $\mathcal O_t$ includes multi-view visual, fingertip tactile, tool force-torque, and audio signals. All signal streams are synchronized using timestamps. Visual, tactile, force-torque, and proprioceptive histories contain $T=16$ consecutive steps ending at $t$. They are aligned to the 30 Hz front-view camera timeline and spanning approximately 0.5 s~\cite{motoda2026self}. Following ManiWAV~\cite{liu2025maniwav}, the audio history is a continuous two-second segment ending at $t$.

We learn a bimanual dexterous manipulation policy $\pi_\phi(\mathrm A_t\mid l,\mathcal O_t,\mathrm P_t)$ from expert demonstrations, conditioned on a natural-language task instruction $l$, multimodal observation history $\mathcal O_t$, and proprioceptive history $\mathrm P_t$. Each action chunk $\mathrm A_t=[a_t,\ldots,a_{t+H-1}]$ contains $H$ consecutive actions. $\mathcal O_t$ comprises multi-view vision, fingertip tactile, tool force-torque, and audio signals. All data are synchronized using timestamps. Visual, tactile, force-torque, and proprioceptive histories contain $T=16$ consecutive steps ending at $t$ on the 30\,Hz front-view camera timeline, spanning approximately 0.5\,s~\cite{motoda2026self}. Following ManiWAV~\cite{liu2025maniwav}, audio is a continuous two-second segment ending at $t$.

\subsection{Architecture Overview.}

As shown in Fig.~\ref{fig:pipeline}, HiMoDex follows two-stage training. In Stage 1 (S1), we pretrain a hierarchical multimodal representation network $\mathcal{F}_{\theta}$ to encode temporally aligned observation histories into unified world-state embeddings. We maintain an online network $\mathcal{F}_{\theta}$ and a target network $\mathcal{F}_{\bar{\theta}}$, whose parameters are updated via an exponential moving average (EMA) of the online parameters. The online network is optimized using three pretraining objectives: cross-representation consistency, signal reconstruction and prediction, and task-state contrastive learning. In Stage 2 (S2), we freeze the EMA target network $\mathcal{F}_{\bar{\theta}}$ from S1 and use it to encode observation histories from downstream tasks into world-state embeddings. A shared world-action transformer uses these embeddings to jointly learn action generation and action-conditioned future world-state prediction.

\subsection{S1: Hierarchical Multimodal Representation.}

In S1, $\mathcal{F}_{\theta}$ encodes visual and non-visual observation histories into visual context embeddings $Z_V$ and interaction embeddings $Z_I$, respectively. A world-state encoder then integrates these embeddings into unified world-state embeddings $Z_W$. This hierarchical representation captures both visual context and physical interactions.

\noindent\textbf{Visual Context Encoding.}
Multi-view visual signals are acquired from one front-view camera and two wrist-mounted cameras. Each RGB frame is resized to $384\times384$ pixels and normalized. Each camera sequence is then independently encoded by a frozen V-JEPA 2.1 ViT-L backbone~\cite{assran2025vjepa2,mur2026vjepa21}. The extracted features $\mathrm{X}_t^{\mathrm{v}}$ are spatially averaged~\cite{musa2026and} and temporally interpolated to $T$ observation steps. A trainable linear layer projects the aligned features to dimension $D$. The visual context encoder then uses $N_{\mathrm{V}}$ learnable visual context queries to integrate these multi-view temporal features~\cite{jaegle2021perceiver}. It consists of two cross-attention blocks followed by four self-attention blocks. The resulting visual context embeddings are $\mathrm{Z}_{\mathrm{V}}\in\mathbb{R}^{T\times N_{\mathrm{V}}\times D}$.
% We then add learnable embeddings for modality identity, camera identity, and temporal position.

\noindent\textbf{Physical Interaction Representation.}
We encode tactile, force-torque, and audio histories with modality-specific Intra-Inter Sensor Transformers and fuse their outputs into interaction embeddings via Gated Interaction Fusion.

\textbf{Non-visual signal preprocessing.} Tactile signals are collected from the five fingertips of each BrainCo Revo2 hand. We normalize the normal force and tangential force magnitude and convert the tangential direction to its sine and cosine components. Force-torque signals are collected from two end-effector sensors, each measuring three force and three torque components. Modality-specific multilayer perceptron (MLP) networks project these measurements into features $\mathrm{X}_t^{\mathrm{tac}}\in\mathbb{R}^{T\times10\times D}$ and $\mathrm{X}_t^{\mathrm{f-t}}\in\mathbb{R}^{T\times2\times D}$. Audio from two hand-mounted microphones is converted to log-mel spectrograms~\cite{liu2025maniwav}. A convolutional projection and cross-attention with learnable queries~\cite{jaegle2021perceiver} produce audio features $\mathrm X_t^{\mathrm{aud}}\in\mathbb R^{T\times2\times D}$ aligned with the $T$ observation steps. These are augmented with learnable embeddings for modality identity, hand identity, and temporal position.
% \textcolor{blue}{add audio preprocessing details or cite} 

\textbf{Intra-Inter Sensor Transformer.} Physical interaction cues are reflected in both the temporal evolution of individual sensor signals and the variation in responses across sensing locations. To jointly model these temporal and spatial dependencies, we design the Intra-Inter Sensor Transformer (I$^2$ST) to learn spatiotemporal representations. Separate I$^2$ST parameters are used for different modalities.

I$^2$ST comprises three sequential modules. The intra-sensor encoding module uses two 1D ConvNeXt~\cite{liu2022convnext} blocks to capture intra-sensor local temporal dynamics. Each block combines a depthwise temporal convolution with a channel-wise MLP. The inter-sensor attention module then applies self-attention across sensors of the same modality at each time step to model relationships among their responses. Finally, the temporal attention module applies self-attention over the full observation window to capture long-range dependencies within each sensor stream. Finally, I$^2$ST produces modality-specific features $\mathrm{Z}_{\mathrm{tac}}$, $\mathrm{Z}_{\mathrm{f-t}}$, and $\mathrm{Z}_{\mathrm{aud}}$.

\textbf{Gated Interaction Fusion.}
Different modalities provide complementary interaction cues. For example, tactile and force-torque signals can indicate that an electric brush is held, while audio can indicate whether it is powered on. We therefore introduce the Gated Interaction Fusion module to integrate these cues into a unified interaction representation. 
  
Algorithm~\ref{alg:gated_interaction_fusion} summarizes the fusion procedure. Shared interaction tokens exchange information with the modality-specific features $\mathrm{Z}_{\mathrm{tac}}$, $\mathrm{Z}_{\mathrm{f-t}}$, and $\mathrm{Z}_{\mathrm{aud}}$ over $L$ rounds. For each modality, a separate module $E_m^{(\ell)}$ updates its features and generates candidate interaction tokens at each observation step. Parameters are separate across modalities and fusion rounds. A softmax gate across available modalities combines these candidates at each time step and interaction-token position. Gate projections are initialized to zero to ensure uniform initial weights.

\begin{algorithm}[t]
\caption{Gated Interaction Fusion}
\label{alg:gated_interaction_fusion}
\begin{algorithmic}

\Statex \textbf{Input:}
Available modalities
$\mathcal M\subseteq\{\mathrm{tac},\mathrm{f-t},\mathrm{aud}\}$;
features $\{\mathrm Z_m\}_{m\in\mathcal M}$;
learnable interaction tokens $\mathrm I\in\mathbb R^{N_{\mathrm I}\times D}$;
fusion depth $L$

\Statex \textbf{Output:}
$\mathrm Z_{\mathrm I}\in\mathbb R^{T\times N_{\mathrm I}\times D}$

\Statex \textbf{Initialization:} $\mathrm Z_m^{(0)}\gets\mathrm Z_m$, $m\in\mathcal M$; broadcast $\mathrm I$ over $T$ steps to obtain $\mathrm I^{(0)}$

\For{$\ell=0,\ldots,L-1$}
    \If{$\mathcal M=\varnothing$}
        \State $\mathrm I^{(\ell+1)}\gets\mathrm I^{(\ell)}$
    \Else
        \For{$m\in\mathcal M$}
            \State $\mathrm U_m\gets E_m^{(\ell)}
            \bigl(\operatorname{Concat}
            (\mathrm Z_m^{(\ell)},\mathrm I^{(\ell)})\bigr)$
            \State $(\mathrm Z_m^{(\ell+1)},\widetilde{\mathrm I}_m)
            \gets\operatorname{Split}(\mathrm U_m)$
            \State $s_m\gets\operatorname{Linear}_m^{(\ell)}
            \bigl(\operatorname{RMSNorm}(\widetilde{\mathrm I}_m)\bigr)$
        \EndFor
        \State $(\alpha_m)_{m\in\mathcal M}
        \gets\operatorname{Softmax}_{\mathcal M}
        \bigl((s_m)_{m\in\mathcal M}\bigr)$
        \State $\displaystyle
        \mathrm I^{(\ell+1)}
        \gets\sum_{m\in\mathcal M}
        \alpha_m\odot\widetilde{\mathrm I}_m$
    \EndIf
\EndFor

\State Flatten $\mathrm I^{(L)}$ into $T\times N_{\mathrm I}$, apply two self-attention layers, and obtain $\mathrm Z_{\mathrm I}$
\end{algorithmic}
\end{algorithm}

\noindent\textbf{World-state Encoder.}
At each observation step, we concatenate $N_{\mathrm W}$ learnable world-state queries with the visual context embeddings $\mathrm Z_{\mathrm V}$ and interaction embeddings $\mathrm Z_{\mathrm I}$. A Transformer module jointly updates these tokens. The outputs at the world-state query positions are then arranged into a sequence of length $T\times N_{\mathrm W}$. A second Transformer module models dependencies across time and query positions, producing world-state embeddings $\mathrm Z_{\mathrm W}\in\mathbb R^{T\times N_{\mathrm W}\times D}$.

\noindent\textbf{Pretraining objectives.} During pretraining, $\mathcal{F}_{\theta}$ processes observation histories with randomly dropped non-visual modalities, while $\mathcal{F}_{\bar{\theta}}$ receives the same histories without masking. We denote the resulting online and target embeddings by $\mathrm{Z}_u$ and $\overline{\mathrm{Z}}_u$, respectively, where $u\in\{\mathrm{V},\mathrm{I},\mathrm{W}\}$. We optimize $\mathcal{F}_{\theta}$ using the following three objectives.

\emph{(1) Cross-representation consistency.} Following latent-space prediction with EMA targets~\cite{assran2023ijepa}, we predict full-observation target embeddings from online embeddings along three directions: $\mathcal{R}=\{\mathrm{W}\rightarrow\mathrm{V}, \mathrm{W}\rightarrow\mathrm{I}, \mathrm{V}\rightarrow\mathrm{I}\}$. The first two encourage $\mathrm{Z}_\mathrm{W}$ to retain visual context and physical interaction information. The third encourages $\mathrm{Z}_\mathrm{V}$ to capture visual cues relevant to physical interactions. Each direction uses an independent predictor $f_{r\rightarrow g}$. At each time step, cross-attention updates $N_g$ learnable queries using $\mathrm{Z}_r$ as keys and values. The resulting $T\times N_g$ tokens are then processed by self-attention~\cite{jaegle2021perceiver}. The prediction matches the shape of its target $\overline{\mathrm{Z}}_g\in\mathbb{R}^{T\times N_g\times D}$. All EMA targets are stop-gradient. In addition, to improve the robustness of world-state representations, we align online world-state embeddings $\mathrm{Z}_{\mathrm{W}}$ with their full-observation EMA targets $\overline{\mathrm{Z}}_{\mathrm{W}}$ under non-visual modality dropout. The combined loss is
\begin{equation}
\label{eq:cross_representation_objective}
\begin{aligned}
\mathcal{L}_{\mathrm{cross}} ={}& \sum_{(r\rightarrow g)\in\mathcal{R}} \lambda_{r\rightarrow g}\, \mathbb{E}\!\left[ \ell_{\mathrm{reg}}\!\left( f_{r\rightarrow g}(\mathrm{Z}_r), \overline{\mathrm{Z}}_g \right) \right] \\ &+ \lambda_{\mathrm{con}}\, \mathbb{E}_{\mathrm{drop}}\!\left[ \ell_{\mathrm{reg}}\!\left( \mathrm{Z}_{\mathrm{W}}, \overline{\mathrm{Z}}_{\mathrm{W}} \right) \right], 
\end{aligned}
\end{equation}
where $\ell_{\mathrm{reg}}$ is the mean-squared error. $\mathbb E_{\mathrm{drop}}$ averages over windows with at least one dropped non-visual modality. We set $\lambda_{\mathrm{W}\rightarrow\mathrm{V}} =\lambda_{\mathrm{W}\rightarrow\mathrm{I}}=0.5$ and $\lambda_{\mathrm{V}\rightarrow\mathrm{I}} =\lambda_{\mathrm{con}}=1.0$.

% $\ell_{\mathrm{reg}}(A,B) =\operatorname{MSE}(\operatorname{LN}(A),\operatorname{LN}(B))$. Layer normalization (LN) is applied to each embedding along the feature dimension, and MSE is averaged over time steps, tokens, and features. $\mathbb{E}$ averages over training windows and sampled masks, whereas $\mathbb{E}_{\mathrm{drop}}$ considers only windows with at least one dropped non-visual modality.

\emph{(2) Signal reconstruction and change prediction.} We reconstruct masked tactile and force-torque measurements from online interaction embeddings $\mathrm{Z}_{\mathrm{I}}$. We randomly select one retained tactile or force-torque modality and mask a contiguous measurement segment. We implement masking by replacing the projected modality features in this segment with a learnable mask token. The online network then encodes the masked observations into $\mathrm{Z}_{\mathrm{I}}$. At each time step, we average the $N_{\mathrm{I}}$ embeddings in $\mathrm{Z}_{\mathrm{I}}$ and use a modality-specific MLP to reconstruct the masked measurements. The prediction targets comprise normalized and smoothed tactile force magnitudes and end-effector force-torque components. The reconstruction loss $\mathcal{L}_{\mathrm{est}}$ is the smooth $L_1$ loss at masked positions. Reconstruction is skipped when neither modality is retained.

For each retained modality $m\in\{\mathrm{tac},\mathrm{f-t}\}$, we also predict the change $\Delta\mathbf y_m$ between the two halves of the $T$-step observation window.
\begin{equation}
\label{eq:signal_change_target}
\Delta\mathbf y_m = \frac{1}{|\mathcal T_2|} \sum_{\tau\in\mathcal T_2}\mathbf y_{m,\tau} - \frac{1}{|\mathcal T_1|} \sum_{\tau\in\mathcal T_1}\mathbf y_{m,\tau},
\end{equation} 
where $\mathbf y_{m,\tau}$ is the target measurement vector. The index sets $\mathcal T_1$ and $\mathcal T_2$ correspond to the first and second halves, with $|\mathcal T_1|=|\mathcal T_2|=T/2$. We average $\mathrm Z_{\mathrm I}$ within each half  and concatenate the two resulting vectors. A separate modality-specific MLP uses this vector to predict $\widehat{\Delta\mathbf{y}}_m$. We compute the change loss $\mathcal{L}_{\Delta}$ from the smooth $L_1$ errors between $\widehat{\Delta\mathbf{y}}_m$ and $\Delta\mathbf{y}_m$. Following~\cite{cheng2026tau}, we assign larger sample weights to windows with greater signal changes when computing this loss. The combined loss is $\mathcal{L}_{\mathrm{sig}} =\lambda_{\mathrm{est}}\mathcal{L}_{\mathrm{est}} +\lambda_{\Delta}\mathcal{L}_{\Delta}$. We set $\lambda_{\mathrm{est}}=\lambda_{\Delta}=0.1$.

\emph{(3) Task-state contrastive learning.} We define task states as manipulation stages within each task and assign a state label to each time step. This objective encourages consistent representations of the same stage across demonstrations while distinguishing different stages. State labels are obtained through an LLM-assisted annotation pipeline based on multimodal cues. These labels supervise the learning of representations that associate each task stage with its characteristic contact patterns.
% We define task states by manipulation stages, such as holding an eraser or wiping a surface. State labels are obtained through an LLM-assisted annotation pipeline and manually spot-checked.

Within each mini-batch, we construct triplets consisting of an anchor, a positive with the same state label, and a negative with a different state label. For each triplet, we apply a triplet-margin loss~\cite{schroff2015facenet} with cosine similarity separately to $\mathrm Z_u$ for $u\in\{\mathrm V,\mathrm I,\mathrm W\}$. The weighted sum of the three losses defines $\mathcal{L}_{\mathrm{state}}$, with each coefficient set to $0.1$.

Overall, the pretraining objective of S1 is the sum of $\mathcal{L}_{\mathrm{cross}}$, $\mathcal{L}_{\mathrm{sig}}$, and $\mathcal{L}_{\mathrm{state}}$.

\begin{figure*}[tbp]
  \centering
  \includegraphics[width=0.95\linewidth]{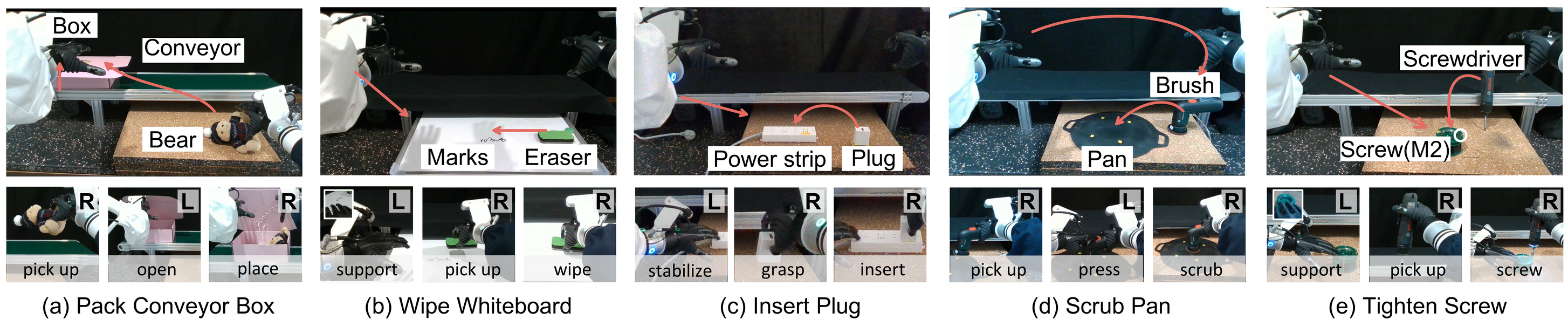}
  \caption{ Overview of the five real-world evaluation tasks. (a) Pack Conveyor Box (Pack), (b) Wipe Whiteboard (Wipe), (c) Insert Plug (Insert), (d) Scrub Pan (Scrub), and (e) Tighten Screw (Tighten). These tasks span diverse manipulation skills. 
  % The lower panels illustrate key manipulation steps, with L and R denoting the left and right hands, respectively.
  }
  \label{fig:task_setting}
\end{figure*}

\begin{figure*}[tbp]
  \centering
\includegraphics[width=0.95\linewidth]{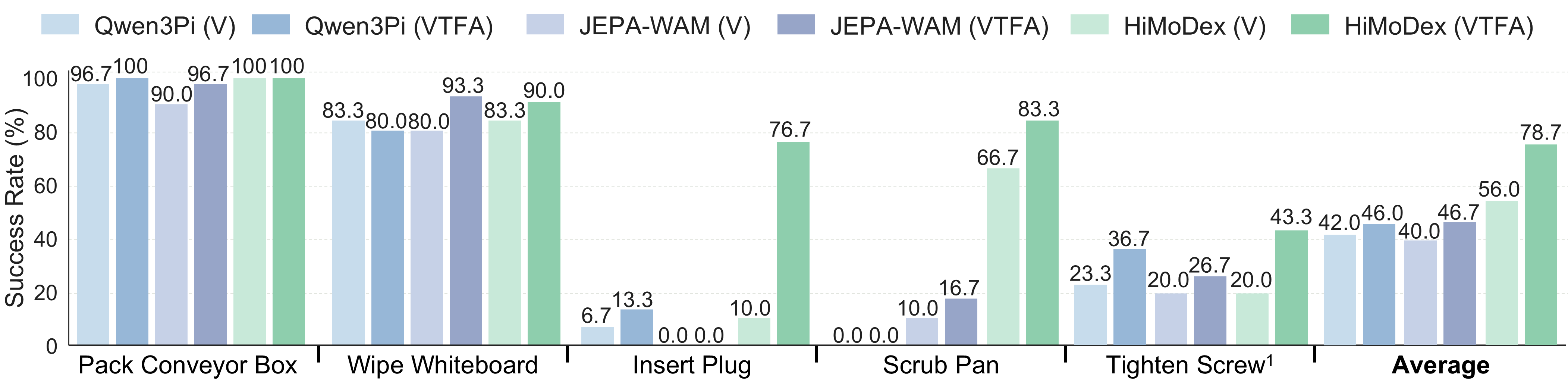}
  \caption{ Performance comparison across five real-world bimanual dexterous manipulation tasks. Each method is evaluated over 30 trials per task, and HiMoDex achieves the highest average success rate. No method completes the particularly challenging Tighten Screw task (as described in Sec.~\ref{subsection:task_setting}). Tighten Screw$^{1}$ indicates the success rate of grasping the screwdriver and bringing it near the screw. The average is computed over the full-task success rates of the first four tasks and the subgoal success rate for Tighten Screw$^{1}$.
  }
  \label{fig:main_results}
\end{figure*}

\subsection{S2: World-Action Policy Learning.}
In S2, we learn a manipulation policy for downstream tasks. We use the frozen EMA network $\mathcal F_{\bar\theta}$ pretrained in S1 to extract world-state embeddings. A shared world-action transformer is jointly trained for action generation and action-conditioned future world-state prediction. 

\noindent\textbf{Shared World-Action Transformer.} During training, it receives five groups of inputs. \emph{1) State embeddings $\mathrm Z_{\mathrm{s}}$.} We concatenate world-state embeddings from the frozen $\mathcal F_{\bar\theta}$ with proprioceptive embeddings to form $\mathrm Z_{\mathrm s}$. Each proprioceptive state in $\mathrm P_t$ contains 12 arm joint positions, 12 end-effector pose components, and 12 hand joint positions. \emph{2) Action queries $\mathrm Q_{\mathrm{a}}$.} The action head uses these $H$ learnable queries to generate $\mathrm A_t\in\mathbb R^{H\times36}$. Each action matches the dimensionality of a proprioceptive state. \emph{3) Ground-truth action embeddings $\mathrm Z_{\mathrm{a}}$.} An MLP encodes the $T$ ground-truth actions aligned with the future prediction window. With temporal position embeddings added, $\mathrm Z_{\mathrm a}$ conditions future world-state prediction. \emph{4) Future world-state queries $\mathrm Q_{\mathrm{f}}$.} These $T \times N_{\mathrm W}$ learnable queries are used to predict future world-state embeddings. Both $\mathrm Z_{\mathrm{a}}$ and $\mathrm Q_{\mathrm{f}}$ are used only during training.
\emph{5) Task instruction $l$.} A frozen UMT5~\cite{chung2023umt5} encoder maps $l$ to instruction embeddings. Only the action queries $\mathrm Q_{\mathrm a}$ cross-attend to these embeddings.

% Each transformer block applies masked self-attention, instruction cross-attention, and a feed-forward network. Following Fast-WAM~\cite{yuan2026fastwam}, the attention mask in Fig.~\ref{fig:pipeline} prevents action queries from accessing training-only information. Attention is bidirectional within each token group.

Each transformer block applies masked self-attention, instruction cross-attention, and a feed-forward network. Following Fast-WAM~\cite{yuan2026fastwam}, we use a structured attention mask (Fig.~\ref{fig:pipeline}). Action queries attend to $\mathrm Z_{\mathrm s}$ and one another, while future world-state queries additionally access $\mathrm Z_{\mathrm a}$. Attention is bidirectional within each token group. This mask prevents ground-truth actions and future prediction features from entering the action generation branch.

\noindent\textbf{Training Objective.} We train the action head with flow matching~\cite{lipman2022flow}, conditioned on the transformer outputs at $\mathrm Q_{\mathrm a}$. Given noisy actions and flow time, the head predicts velocities along linear paths from Gaussian noise to expert action chunks $\mathrm A_t$. The action loss $\mathcal L_{\mathrm{FM}}$ is the mean-squared error between predicted and target velocities.

For future world-state prediction, an MLP maps the transformer outputs at $\mathrm Q_{\mathrm f}$ to predicted embeddings. The frozen EMA network encodes the future observation window to provide the targets. The world prediction loss $\mathcal L_{\mathrm W}$ is the mean-squared error between predicted and target embeddings.  Both losses train the shared transformer through $\mathcal L_{\mathrm{policy}} =\mathcal L_{\mathrm{FM}}+\lambda_{\mathrm W}\mathcal L_{\mathrm W}$, with $\lambda_{\mathrm W}=1.0$. At inference, we retain only the action generation branch.

% An MLP maps the outputs at $\mathrm Q_{\mathrm{f}}$ to future world-state embeddings. Targets are obtained by applying the frozen EMA network from Stage 1 to the future observation window from $t+1$ to $t+T$. The world prediction loss $\mathcal L_{\mathrm W}$ is the mean-squared error between predicted and target embeddings. Both losses update the shared transformer through the joint objective $\mathcal L_{\mathrm{policy}} =\mathcal L_{\mathrm{FM}}+\lambda_{\mathrm W}\mathcal L_{\mathrm W}$. At inference, we retain only the action generation branch to improve efficiency.

More details of S1 and S2 are provided in the Appendix.

\section{Experiments}
\label{sec:experiments}

\begin{table*}[!t]
\centering
\caption{Ablation results over 10 trials per configuration and task. Success rates (SR, \%). F1-F3 report task-specific failure-type counts (0-10). Darker shades indicate higher counts. S1 and S2 in the group labels denote ablations targeting Stage~1 and Stage~2, respectively.
% Bold denotes the highest SR or average in each column, including ties.
}
\label{tab:ablations}

\begingroup
\setlength{\tabcolsep}{2.8pt}
\renewcommand{\arraystretch}{1.}
\resizebox{0.9\textwidth}{!}{%
% 21 columns: 3 shared labels + task blocks of 3,4,3,4,3 + average.
% 2pt between SR and failures; 0pt within failure blocks.
% Light gray rules sit within the original 5pt / 6pt boundary gaps.
\begin{tabular}{@{}lcl@{\AblationConfigSep}
  S{10}@{\hspace{2pt}}A@{}A@{\AblationTaskSep}
  S{9}@{\hspace{2pt}}A@{}A@{}A@{\AblationTaskSep}
  S{7}@{\hspace{2pt}}A@{}A@{\AblationTaskSep}
  S{8}@{\hspace{2pt}}A@{}A@{}A@{\AblationTaskSep}
  S{5}@{\hspace{2pt}}A@{}A@{\AblationTaskSep}
  B{74}@{}}
\toprule
\multirow{2}{*}{Group} &
\multirow{2}{*}{ID} &
\multirow{2}{*}{Configuration} &
\multicolumn{3}{@{}c@{\AblationTaskSep}}{Pack} &
\multicolumn{4}{@{}c@{\AblationTaskSep}}{Wipe} &
\multicolumn{3}{@{}c@{\AblationTaskSep}}{Insert} &
\multicolumn{4}{@{}c@{\AblationTaskSep}}{Scrub} &
\multicolumn{3}{@{}c@{\AblationTaskSep}}{Tighten$^1$} &
\multirow{2}{*}{Avg.$\uparrow$} \\
\cmidrule(l{0pt}r{6pt}){4-6}\cmidrule(l{0pt}r{6pt}){7-10}
\cmidrule(l{0pt}r{6pt}){11-13}\cmidrule(l{0pt}r{6pt}){14-17}
\cmidrule(l{0pt}r{6pt}){18-20}
& & & \HeaderTwo & \HeaderThree & \HeaderTwo &
\HeaderThree & \HeaderTwo & \\
\midrule

% I: Full model.
I & 01 & HiMoDex(VTFA)
& \TaskTwo{10}{0}{0} & \TaskThree{8}{0}{0}{2}
& \TaskTwo{7}{0}{3} & \TaskThree{8}{2}{0}{0}
& \TaskTwo{4}{6}{0} & \AverageCell{74} \\
\midrule

% II: Stage-1 training objectives.
\multirow{3}{*}{II-S1} & 02 & w/o $\mathcal{L}_{\mathrm{sig}}$
& \TaskTwo{10}{0}{0} & \TaskThree{9}{0}{0}{1}
& \TaskTwo{1}{6}{3} & \TaskThree{0}{0}{10}{0}
& \TaskTwo{2}{8}{0} & \AverageCell{44} \\
& 03 & w/o $\mathcal{L}_{\mathrm{state}}$
& \TaskTwo{10}{0}{0} & \TaskThree{8}{0}{0}{2}
& \TaskTwo{0}{0}{10} & \TaskThree{1}{9}{0}{0}
& \TaskTwo{3}{5}{2} & \AverageCell{44} \\
& 04 & w/o $(\mathcal{L}_{\mathrm{sig}}+\mathcal{L}_{\mathrm{state}})$
& \TaskTwo{7}{1}{2} & \TaskThree{7}{0}{0}{3}
& \TaskTwo{3}{0}{7} & \TaskThree{1}{0}{0}{9}
& \TaskTwo{1}{5}{4} & \AverageCell{38} \\
\midrule

% III: Stage-1 fusion designs.
\multirow{2}{*}{III-S1} & 05 & Flat
& \TaskTwo{9}{0}{1} & \TaskThree{7}{2}{0}{1}
& \TaskTwo{4}{2}{4} & \TaskThree{7}{3}{0}{0}
& \TaskTwo{2}{5}{3} & \AverageCell{58} \\
& 06 & H-concat
& \TaskTwo{9}{0}{1} & \TaskThree{9}{0}{0}{1}
& \TaskTwo{0}{6}{4} & \TaskThree{4}{3}{3}{0}
& \TaskTwo{2}{8}{0} & \AverageCell{48} \\

% & 07 & H-cross attention
% & \TaskTwo{9}{0}{1} & \TaskThree{9}{0}{0}{1}
% & \TaskTwo{1}{0}{9} & \TaskThree{2}{0}{8}{0}
% & \TaskTwo{4}{6}{0} & \AverageCell{50} \\

\midrule

% IV: Stage-1 interaction gates.
IV-S1 & 08 & w/o gates
& \TaskTwo{9}{0}{1} & \TaskThree{8}{0}{0}{2}
& \TaskTwo{2}{0}{8} & \TaskThree{7}{3}{0}{0}
& \TaskTwo{4}{6}{0} & \AverageCell{60} \\
\midrule

% V: Stage-2 observation modalities.
\multirow{3}{*}{V-S2} & 09 & V
& \TaskTwo{10}{0}{0} & \TaskThree{7}{0}{0}{3}
& \TaskTwo{1}{0}{9} & \TaskThree{7}{2}{0}{1}
& \TaskTwo{2}{8}{0} & \AverageCell{54} \\
& 10 & VT
& \TaskTwo{10}{0}{0} & \TaskThree{7}{0}{0}{3}
& \TaskTwo{7}{0}{3} & \TaskThree{7}{2}{0}{1}
& \TaskTwo{4}{6}{0} & \AverageCell{70} \\
& 11 & VTF
& \TaskTwo{10}{0}{0} & \TaskThree{9}{0}{0}{1}
& \TaskTwo{6}{0}{4} & \TaskThree{6}{4}{0}{0}
& \TaskTwo{5}{5}{0} & \AverageCell{72} \\
\midrule

% VI: Stage-2 future world-state supervision.
VI-S2 & 12 & w/o $\mathcal L_{\mathrm W}$
& \TaskTwo{10}{0}{0} & \TaskThree{8}{0}{0}{2}
& \TaskTwo{2}{0}{8} & \TaskThree{6}{0}{0}{4}
& \TaskTwo{2}{8}{0} & \AverageCell{56} \\
\midrule

% VII: Training strategy.
VII & 13 & End-to-end
& \TaskTwo{10}{0}{0} & \TaskThree{5}{0}{3}{2}
& \TaskTwo{3}{0}{7} & \TaskThree{4}{4}{0}{2}
& \TaskTwo{4}{6}{0} & \AverageCell{52} \\
\bottomrule
\end{tabular}%
}
\endgroup

\vspace{2pt}
\begin{minipage}{\textwidth}
\footnotesize
\textbf{Failure count:}\quad
\begin{tikzpicture}[baseline=0pt,x=1cm,y=1cm]
  \foreach \n in {0,...,10} {
    \pgfmathtruncatemacro{\tone}{10*\n}
    \fill[FailureBlue!\tone!FailureZero]
      ({0.32*\n},0) rectangle ({0.32*(\n+1)+0.004},0.18);
    \node[below,font=\scriptsize,inner sep=1pt]
      at ({0.32*(\n+0.5)},-0.025) {\n};
  }
\end{tikzpicture}

\smallskip
\textbf{Failure types.}
\textit{Pack:} F1, insecure object grasp; F2, failure to grasp the object.
\textit{Wipe:} F1, failure to grasp the eraser; F2, insecure eraser grasp; F3, residual marks after wiping.
\textit{Insert:} F1, failure to grasp the plug; F2, failed insertion.
\textit{Scrub:} F1, insufficient pressing force to activate the switch; F2, off-target pressing that fails to activate the switch; F3, insufficient scrubbing strokes.
\textit{Tighten:} F1: failure to grasp the screwdriver; F2: failure to approach the screw.
% \smallskip
% \textbf{Counting.}
% Each failed trial is assigned to a single failure category. Zero counts are unlabeled and use the lightest shade.
\end{minipage}
\end{table*}

\subsection{Experimental Setups}
\label{subsection:task_setting}

\noindent\textbf{Stage 1: Multimodal Pretraining Data.} We collected 1,187,768 time steps of synchronized multimodal observations using our robotic platform. These yielded 291,578 aligned observation windows of $T=16$ steps. We used 287,718 windows for pretraining and reserved 3,860 for representation evaluation. More details are in the Appendix.
% \hl{Dataset and evaluation details are provided in the appendix.}
% We assess state discrimination and cross-episode consistency for visual context, interaction, and world-state embeddings. We also evaluate the robustness of world-state embeddings to missing non-visual modalities.

\noindent\textbf{Stage 2: Downstream Tasks.} We collected 770 expert demonstrations across five bimanual dexterous manipulation tasks to train the WAM policy. These tasks assess a range of manipulation capabilities. We varied initial object positions and orientations across demonstrations
to increase data diversity. Despite some scenario overlap with S1, S2 demonstrations were collected independently with different environmental conditions. They were excluded from S1 pretraining. Fig.~\ref{fig:task_setting} shows the five task settings.

% Although some scenarios overlap with those in S1, the Stage 2 demonstrations were collected independently. The two datasets differ in environmental conditions, robot and object placement, and manipulated objects. No Stage 2 demonstrations were used for Stage 1 pretraining. Fig.~\ref{fig:task_setting} shows the five task settings.

\textit{Task 1: Pack Conveyor Box. }The right hand grasps a toy bear while the left hand opens the lid of a box on a conveyor belt. The right hand then places the toy inside. This task evaluates basic bimanual coordination.

\textit{Task 2: Wipe Whiteboard.} The robot grasps an eraser and wipes marks from the whiteboard. This task evaluates the ability to sustain controlled contact.

\textit{Task 3: Insert Plug.} The left hand stabilizes a power strip. The right hand grasps a power plug and inserts it into the designated slot. This task evaluates the ability to perform precise, contact-rich manipulation.

\textit{Task 4: Scrub Pan.} The right hand grasps an electric brush and the left hand presses its activation button. After activation, the brush makes a sound. The right hand then scrubs the bottom of the pan. This task evaluates coordinated bimanual tool use.

\textit{Task 5: Tighten Screw.} We introduce this task as the most challenging evaluation setting. It probes the limits of the model's capabilities. First, the front-view camera is 0.6\,m from the object, making the M2 screw difficult to observe clearly, as shown in Fig.~\ref{fig:task_setting}(e). The left arm must therefore stabilize the workpiece while positioning its wrist-mounted camera for a closer view. Second, grasping is more challenging than in \textit{Scrub Pan}, as accidental contact can topple the unsecured, upright electric screwdriver. Third, the right hand must precisely align the screwdriver with the screw head and tighten the 5\,mm-long screw. This near-instantaneous tightening phase is difficult to learn.

% Second, tool grasping is more demanding than in \textit{Scrub Pan}. The electric screwdriver stands upright without being secured. Accidental contact can topple it and cause grasp failure. 
% Together, these requirements demand more precise bimanual coordination than Insert Plug.

% \textbf{Evaluation.}
% We evaluate policy performance primarily using task success rates across all tasks. For the main comparisons, we evaluate each method over 30 trials per task. For ablations, we evaluate each variant over 10 trials per task.
% We evaluate policies by task success rate, using 30 trials per task for main comparisons and 10 for ablations.

% \textbf{Baselines.} We compare HiMoDex with Qwen3Pi and JEPA-WAM~\cite{lin2026jepawam} as representative VLA and WAM baselines, respectively. Qwen3Pi is implemented using StarVLA~\cite{community2026starvla} with a Qwen3-VL backbone and a flow-matching action head. Its design incorporates choices from $\pi_0$~\cite{black2025pi0} and $\pi_{0.5}$~\cite{physicalintelligence2025pi05}.  For each framework, we evaluate vision-only (V) and full-modality (VTFA) variants. The latter uses visual, tactile, force-torque, and audio inputs.
% For each framework, we evaluate a vision-only variant (V) and a multimodal variant (VTFA). The latter incorporates visual, tactile, force-torque, and audio.
% This yields four baselines: Qwen3Pi-V, Qwen3Pi-VTFA, JEPA-WAM-V, and JEPA-WAM-VTFA.

\textbf{Evaluation Metrics and Baselines.} We evaluate policies by task success rate, using 30 trials per task for main comparisons and 10 for ablations. Detailed success criteria are provided in the Appindex. We compare HiMoDex with Qwen3Pi and JEPA-WAM~\cite{lin2026jepawam} as representative VLA and WAM baselines, respectively. Qwen3Pi is implemented using StarVLA~\cite{community2026starvla} with a Qwen3-VL~\cite{bai2025qwen3} backbone and a flow-matching action head. Its design incorporates choices from $\pi_0$~\cite{black2025pi0} and $\pi_{0.5}$~\cite{physicalintelligence2025pi05}.  For each framework, we evaluate vision-only (V) and full-modality (VTFA) variants. The latter uses visual, tactile, force-torque, and audio inputs.

% \textbf{Evaluation Metrics and Protocols.} We evaluate policies by task success rate, using 30 trials per task for main comparisons and 10 for ablations. 

\subsection{Main Results}
% Fig.~\ref{fig:main_results} summarizes performance across the five tasks. Tighten Screw is designed to be particularly challenging. No evaluated method completes the full task. We therefore report the success rate of grasping the screwdriver and bringing it near the screw. This intermediate stage is labeled Tighten Screw$^{1}$ in the Fig.~\ref{fig:main_results}. 
% The reported average combines this intermediate-stage success rate with the full-task success rates of the other four tasks.
As shown in Fig.~\ref{fig:main_results}, HiMoDex (VTFA) achieves an average success rate of 78.7\% across five tasks, outperforming all baseline configurations. It achieves absolute gains of 32.7\% and 32.0\% in success rate over the VTFA variants of Qwen3Pi and JEPA-WAM, respectively. The advantage is pronounced on Insert Plug and Scrub Pan. HiMoDex achieves success rates of 76.7\% and 83.3\%, whereas the strongest baseline results are only 13.3\% and 16.7\%, respectively. These results highlight its effectiveness in manipulation tasks involving precise alignment, sustained contact, and tool use. From V to VTFA, HiMoDex's average success rate rises from 56.0\% to 78.7\%. This gain of 22.7\% exceeds the corresponding gains of 4.0\% for Qwen3Pi and 6.7\% for JEPA-WAM. HiMoDex improves substantially on Insert Plug, Scrub Pan, and Tighten Screw$^1$. These differences suggest that the benefits of additional modalities depend on how effectively a model uses them.

% Gains from additional modalities vary across tasks. 
On Pack Conveyor Box, all variants achieve at least 90.0\% success. Additional modalities provide modest gains, suggesting that visual observations support strong performance on this task. On the most challenging task, Tighten Screw$^{1}$, HiMoDex (VTFA) achieves 43.3\% success, compared with 36.7\% for Qwen3Pi (VTFA). Although the performance gap is modest, HiMoDex uses only 327.62M trainable parameters in S2, approximately one-eighth of Qwen3Pi's 2606.04M. 
% All methods remain below 50\% success on this task. This highlights challenges in precise tool use and the considerable room for improvement.

\begin{figure}[tbp]
  \centering
  \includegraphics[width=0.9\linewidth]{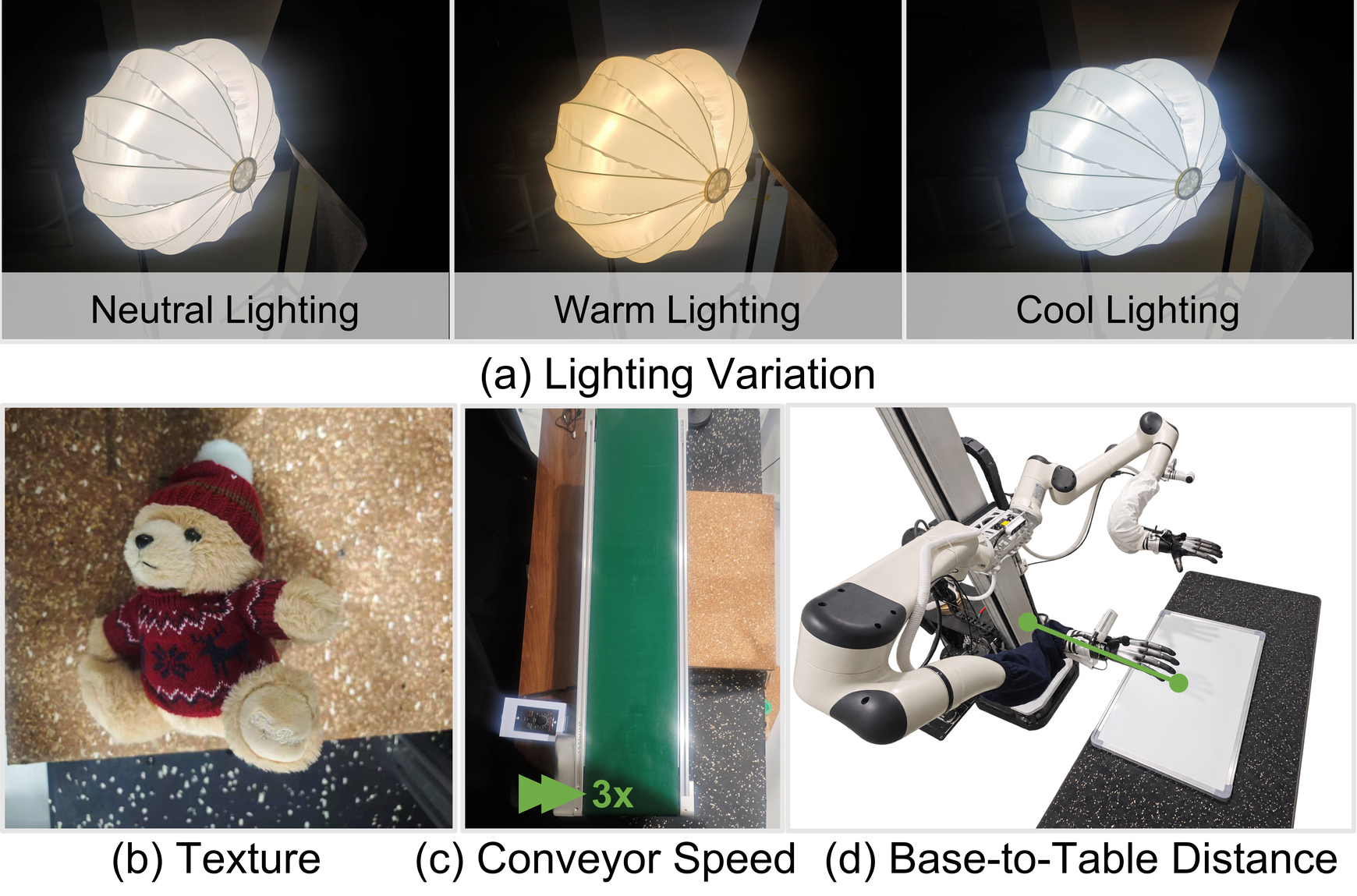}
  \caption{ Settings for generalization evaluation. (a) different lighting conditions, (b) changes in object texture, (c) increased conveyor speed, and (d) variations in robot base-to-table distance.
  }
  \label{fig:generalization_setting}
\end{figure}

\textbf{Model Generalization} We evaluate lighting, texture, and conveyor-speed variations on Pack Conveyor Box (Fig.~\ref{fig:generalization_setting}(a)--(c)) and robot base-to-table distance variations on Wipe Whiteboard (Fig.~\ref{fig:generalization_setting}(d)). As shown in Table~\ref{tab:generalization}, HiMoDex achieves the highest average success rate of 87.5\%, outperforming JEPA-WAM and Qwen3Pi by 7.5 and 10.0 percentage points, respectively. Its high success under lighting and texture changes suggests robustness to visual appearance variations. Under physical variations, HiMoDex outperforms both baselines at altered base-to-table distances and matches the strongest baseline at increased conveyor speed. These results support its ability to handle visual changes while accommodating variations in object motion and workspace geometry.

\subsection{Ablation Studies and Analysis}

Table~\ref{tab:ablations} presents ablation results. Each configuration is evaluated over 10 trials per task. It also reports task-specific failure types and their frequencies. For a fair comparison, we independently reevaluate HiMoDex (VTFA) and HiMoDex (V) using the same checkpoints as in the main evaluation. It obtains average success rates close to those in Fig.~\ref{fig:main_results}.

% Their results therefore differ from those reported in Fig.~\ref{fig:main_results}, but the average success rates remain close across the two evaluations.

\textbf{Effect of S1 Training Objectives.} Group II-S1 examines the contributions of $\mathcal{L}_{\mathrm{sig}}$ and $\mathcal{L}_{\mathrm{state}}$. Removing either objective reduces average success from 74\% to 44\%, while removing both further lowers it to 38\% (02-04). These results highlight the importance of both objectives for downstream policy learning. The failure statistics provide further insight into their respective roles. Without $\mathcal{L}_{\mathrm{sig}}$ (02) increases grasping failures on Insert Plug (F1) from zero to six and off-target pressing failures on Scrub Pan (F2) from zero to ten.
These positioning errors suggest that $\mathcal{L}_{\mathrm{sig}}$ helps learn representations supporting accurate spatial targeting.
In contrast, removing $\mathcal{L}_{\mathrm{state}}$ (03) increases insertion failures on Insert Plug (F2) from three to ten and activation failures caused by insufficient pressing force on Scrub Pan (F1) from two to nine.
This pattern suggests that $\mathcal{L}_{\mathrm{state}}$ helps the policy use contact feedback to adjust its actions. 
% Since task-stage annotations are informed by tactile and force-torque cues, $\mathcal{L}_{\mathrm{state}}$ encourages the representations to associate task stages with characteristic contact and force patterns. These representations can help the downstream policy identify the current interaction stage and adjust its actions according to the contact conditions associated with that stage.

\textbf{Effect of Hierarchical Fusion.} Group III-S1 compares two S1 fusion variants. Flat (05) bypasses the visual context and interaction encoders. It processes modality features and learnable world queries through self-attention to obtain world-state embeddings. H-concat (06) concatenates the visual context and interaction embeddings to form the world representation. Flat reduces average success from 74\% to 58\% and underperforms the full model on all tasks. It suggests a benefit from forming intermediate representations before constructing world-state embeddings. However, retaining these embeddings alone is still insufficient. H-concat achieves 48\% average success, with the largest deficits on Insert Plug and Scrub Pan. On Insert Plug, plug-lifting failures increase from zero to six. These results favor the complete hierarchical design, which combines intermediate representations with a learned world-state encoder.

\textbf{Modality Contributions.} Groups I and V-S2 examine how different modality combinations affect policy performance. As tactile, force, and audio inputs are progressively introduced, the average success rate steadily increases, demonstrating the benefits of multimodal observations. However, additional modalities do not benefit every task. VTF outperforms VTFA on Wipe and Tighten$^1$ (11 vs. 01). The failure breakdown further suggests that tactile input reduces insertion failures on Insert (09 vs. 10), force reduces residual marks on Wipe (10 vs. 11), and audio reduces switch-activation failures on Scrub (11 vs. 01).

% For example, VTF outperforms VTFA on Wipe Whiteboard and Tighten Screw$^1$.
% The failure breakdown further suggests complementary contributions(直接说The failure breakdown further suggests 然后就写后面的): tactile input reduces insertion failures for Insert(09 vs. 10), force reduces residual marks for Wipe(10 vs. 11), and audio reduces switch-activation failures for Scrub(11 vs. 01).

% Specifically, adding tactile input (09 vs. 10) improves the success rate on Insert Plug from 10\% to 70\%, highlighting its importance for precise contact. Adding force input (10 vs. 11) improves Wipe Whiteboard from 70\% to 90\%, with failures due to residual marks decreasing from three to one. Adding audio improves Scrub Pan from 60\% to 80\%, while reducing switch activation failures from four to two (11 vs. 01), suggesting a benefit from auditory feedback during tool activation.

% These results suggest complementary contributions from the three modalities at different stages of physical interaction.

% These gains are task-dependent: VTF outperforms VTFA on Wipe Whiteboard and Tighten Screw, while VTFA achieves the highest average success across the five tasks.

\textbf{Additional Design Choices.}
Replacing the learnable gates in Algorithm~\ref{alg:gated_interaction_fusion} with equal-weight averaging (08) reduces average success from 74\% to 60\% (01 vs. 08). This supports the benefit of adaptively weighting modality contributions. In addition, removing $\mathcal{L}_{\mathrm{W}}$ (12) during S2 reduces average success to 56\%. This supports the value of future-prediction supervision for policy learning, consistent with the findings of Fast-WAM~\cite{yuan2026fastwam}. Finally, we evaluate an end-to-end variant (13) that jointly trains the representation network and policy in a single stage. This variant retains all representation and policy objectives but achieves only 52\% average success. This supports pretraining world-state representations before optimizing the policy.

% The Appendix includes model parameter counts, inference efficiency, and additional experiments and details.
The Appendix provides more results and details.

\begin{table}[t]
    \centering
    \caption{Comparison of generalization performance (\%). The four conditions correspond to Fig.~\ref{fig:generalization_setting}  (a)-(d).}
    \label{tab:generalization}
    \normalsize
    \setlength{\tabcolsep}{2pt}
    \renewcommand{\arraystretch}{1.15}
    \begin{adjustbox}{max width=0.95\columnwidth}
        \begin{tabular}{@{}lccccc@{}}
            \toprule
            Method & Lighting & Texture & Speed & Distance & Avg.$\uparrow$ \\
            \midrule
            Qwen3Pi (VTFA)  & 80 & 100 & 70 & 60 & 77.5 \\
            JEPA-WAM (VTFA) & 80 & 100 & 60 & 80 & 80.0 \\
            HiMoDex (VTFA)  & 90 & 100 & 70 & 90 & \textbf{87.5} \\
            \bottomrule
        \end{tabular}
    \end{adjustbox}
\end{table}

% \begin{table}[t]
%     \centering
%     \caption{Generalization success rates (\%). All methods use VTFA inputs. The four conditions correspond to Fig.~\ref{fig:generalization_setting}(a)-(d).
%     }
%     \label{tab:generalization}
%     \normalsize
%     \setlength{\tabcolsep}{2pt}
%     \renewcommand{\arraystretch}{1.15}
%     \begin{tabular*}{0.9\columnwidth}{@{\extracolsep{\fill}}lccccc@{}}
%         \toprule
%         Method & Lighting & Texture & Speed & Distance & Avg.$\uparrow$ \\
%         \midrule
%         Qwen3Pi (VTFA) & 80 & 100 & 70 & 60 & 77.5 \\
%         JEPA-WAM (VTFA) & 80 & 100 & 60 & 80 & 80.0 \\
%         HiMoDex (VTFA) & 90 & 100 & 70 & 90 & \textbf{87.5} \\
%         \bottomrule
%     \end{tabular*}
% \end{table}
\section{Conclusion}
\label{sec:conclusion}

We propose HiMoDex, a multimodal world-action policy for bimanual dexterous manipulation, learned through two-stage training. In the first stage, we learn hierarchical world-state representations that integrate visual context with tactile, force-torque, and audio interaction cues. In the second stage, we freeze the pretrained encoders and train a shared world-action transformer for action generation and action-conditioned future-state prediction, retaining only action generation at deployment. Experiments across five real-world task settings show that HiMoDex achieves 78.7\% average success, exceeding the strongest baseline by 32.0 percentage points. Ablations validate the benefits of hierarchical multimodal representations and two-stage training. We also contribute a multimodal dataset with 1.19 million synchronized time steps and an expert demonstration dataset for diverse bimanual dexterous manipulation tasks.

\section*{LLM Usage Statement}
During manuscript preparation, we used Codex to assist with language polishing, grammar correction, and suggestions for improving the clarity, organization, and coherence of the manuscript text. All resulting revisions were reviewed by the authors, who take full responsibility for the accuracy and integrity of the final manuscript.
{
    \small
    \bibliographystyle{ieeenat_fullname}
    \bibliography{main}

@String(CVPR= {IEEE Conf. Comput. Vis. Pattern Recog.})

@String(CVPR  = {CVPR})

@inproceedings{shaw2025bidex,
  title = {Bimanual Dexterity for Complex Tasks},
  author = {Shaw, Kenneth and Li, Yulong and Yang, Jiahui and Srirama, Mohan Kumar and Liu, Ray and Xiong, Haoyu and Mendonca, Russell and Pathak, Deepak},
  booktitle = {Proceedings of the Conference on Robot Learning},
  pages = {5166--5183},
  year = {2025},
  volume = {270}
}

@inproceedings{black2025pi0,
  title = {{$\pi_0$}: A Vision-Language-Action Flow Model for General Robot Control},
  author = {Black, Kevin and Brown, Noah and Driess, Danny and Esmail, Adnan and Equi, Michael Robert and Finn, Chelsea and Fusai, Niccolo and Groom, Lachy and Hausman, Karol and Ichter, Brian and Jakubczak, Szymon and Jones, Tim and Ke, Liyiming and Levine, Sergey and Li-Bell, Adrian and Mothukuri, Mohith and Nair, Suraj and Pertsch, Karl and Shi, Lucy Xiaoyang and Smith, Laura and Tanner, James and Vuong, Quan and Walling, Anna and Wang, Haohuan and Zhilinsky, Ury},
  booktitle = {Proceedings of Robotics: Science and Systems},
  year = {2025}
}

@inproceedings{physicalintelligence2025pi05,
  title = {$\pi_{0.5}$: a Vision-Language-Action Model with Open-World Generalization},
  author = {Black, Kevin and Brown, Noah and Darpinian, James and Dhabalia, Karan and Driess, Danny and Esmail, Adnan and Equi, Michael Robert and Finn, Chelsea and Fusai, Niccolo and Galliker, Manuel Y. and Ghosh, Dibya and Groom, Lachy and Hausman, Karol and ichter, brian and Jakubczak, Szymon and Jones, Tim and Ke, Liyiming and LeBlanc, Devin and Levine, Sergey and Li-Bell, Adrian and Mothukuri, Mohith and Nair, Suraj and Pertsch, Karl and Ren, Allen Z. and Shi, Lucy Xiaoyang and Smith, Laura and Springenberg, Jost Tobias and Stachowicz, Kyle and Tanner, James and Vuong, Quan and Walke, Homer and Walling, Anna and Wang, Haohuan and Yu, Lili and Zhilinsky, Ury},
  booktitle = {Proceedings of the Conference on Robot Learning},
  pages = {17--40},
  year = {2025},
  volume = {305}
}

@inproceedings{kim2025openvla,
  title = {OpenVLA: An Open-Source Vision-Language-Action Model},
  author = {Kim, Moo Jin and Pertsch, Karl and Karamcheti, Siddharth and Xiao, Ted and Balakrishna, Ashwin and Nair, Suraj and Rafailov, Rafael and Foster, Ethan P and Sanketi, Pannag R and Vuong, Quan and Kollar, Thomas and Burchfiel, Benjamin and Tedrake, Russ and Sadigh, Dorsa and Levine, Sergey and Liang, Percy and Finn, Chelsea},
  booktitle = {Proceedings of the Conference on Robot Learning},
  pages = {2679--2713},
  year = {2025},
  volume =  {270}
}

@article{yuan2026fastwam,
  title={Fast-wam: Do world action models need test-time future imagination?},
  author={Yuan, Tianyuan and Dong, Zibin and Liu, Yicheng and Zhao, Hang},
  journal={arXiv preprint arXiv:2603.16666},
  year={2026}
}

@article{lin2026jepawam,
  title={JEPA-WAM: Learning vision-language-action policies with joint-embedding world modeling},
  author={Lin, Yihan and He, Jiawei and Bao, Shifeng and Zhao, Chen and Li, Yang and Wang, Xiaobo and Wang, Yan and Chi, Cheng and Zhang, Jing},
  journal={arXiv preprint arXiv:2608.09381},
  year={2026}
}

@inproceedings{lin2025hato,
  title={Learning visuotactile skills with two multifingered hands},
  author={Lin, Toru and Zhang, Yu and Li, Qiyang and Qi, Haozhi and Yi, Brent and Levine, Sergey and Malik, Jitendra},
  booktitle={IEEE International Conference on Robotics and Automation (ICRA)},
  pages={5637--5643},
  year={2025}
}

@article{luo2026beingh05,
  title={Being-h0.5: Scaling human-centric robot learning for cross-embodiment generalization},
  author={Luo, Hao and Wang, Ye and Zhang, Wanpeng and Zheng, Sipeng and Xi, Ziheng and Xu, Chaoyi and Xu, Haiweng and Yuan, Haoqi and Zhang, Chi and Wang, Yiqing and others},
  journal={arXiv preprint arXiv:2601.12993},
  year={2026}
}

@article{yuan2026dextacwam,
  title={DexTacWAM: A Visuo-Tactile World-Action Model for Dexterous Manipulation},
  author={Yuan, Haoran and Wang, Zekai and Shao, Boning and Lu, Haoran and Darrell, Trevor and Lourentzou, Ismini and Zhan, Wei},
  journal={arXiv preprint arXiv:2609.24976},
  year={2026}
}

@article{yu2025forcevla,
  title={Forcevla: Enhancing vla models with a force-aware moe for contact-rich manipulation},
  author={Yu, Jiawen and Liu, Hairuo and Yu, Qiaojun and Ren, Jieji and Hao, Ce and Ding, Haitong and Huang, Guangyu and Huang, Guofan and Song, Yan and Cai, Panpan and others},
  journal={Advances in Neural Information Processing Systems},
  volume={38},
  pages={93409--93439},
  year={2025}
}

@inproceedings{liu2025maniwav,
  title = {ManiWAV: Learning Robot Manipulation from In-the-Wild Audio-Visual Data},
  author = {Liu, Zeyi and Chi, Cheng and Cousineau, Eric and Kuppuswamy, Naveen and Burchfiel, Benjamin and Song, Shuran},
  booktitle = {Proceedings of the Conference on Robot Learning},
  pages = {947--962},
  year = {2025},
  volume = {270}
}

@inproceedings{li2023seehearfeel,
  title = {See, Hear, and Feel: Smart Sensory Fusion for Robotic Manipulation},
  author = {Li, Hao and Zhang, Yizhi and Zhu, Junzhe and Wang, Shaoxiong and Lee, Michelle A and Xu, Huazhe and Adelson, Edward and Fei-Fei, Li and Gao, Ruohan and Wu, Jiajun},
  booktitle = {Proceedings of the Conference on Robot Learning},
  pages = {1368--1378},
  year = {2023},
  volume = {205}
}

@article{kang2025cliprt,
  title={Clip-rt: Learning language-conditioned robotic policies from natural language supervision},
  author={Kang, Gi-Cheon and Kim, Junghyun and Shim, Kyuhwan and Lee, Jun Ki and Zhang, Byoung-Tak},
  journal={arXiv preprint arXiv:2411.00508},
  year={2024}
}

@article{feng2026anytouch2,
  title={Anytouch 2: General optical tactile representation learning for dynamic tactile perception},
  author={Feng, Ruoxuan and Zhou, Yuxuan and Mei, Siyu and Zhou, Dongzhan and Wang, Pengwei and Cui, Shaowei and Fang, Bin and Yao, Guocai and Hu, Di},
  journal={arXiv preprint arXiv:2602.09617},
  year={2026}
}

@article{ma2026fgcltp,
  title={Fg-cltp: Fine-grained contrastive language tactile pretraining for robotic manipulation},
  author={Ma, Wenxuan and Zhang, Chaofan and Cai, Yinghao and Yao, Guocai and Cui, Shaowei and Wang, Shuo},
  journal={IEEE Robotics and Automation Letters},
  year={2026},
  publisher={IEEE}
}

@article{yang2026sota,
  title={SoTa: Soft Tactile Skins for Dexterous Manipulation},
  author={Yang, Jingyun and Shi, Baiyu and Yu, Timothy and Liu, Haitian and Longhini, Alberta and Wang, Weichen and Antonova, Rika and Bao, Zhenan and Bohg, Jeannette},
  journal={arXiv preprint arXiv:2610.02338},
  year={2026}
}

@inproceedings{jones2025fuse,
  title={Beyond sight: Finetuning generalist robot policies with heterogeneous sensors via language grounding},
  author={Jones, Joshua and Mees, Oier and Sferrazza, Carmelo and Stachowicz, Kyle and Abbeel, Pieter and Levine, Sergey},
  booktitle={2025 IEEE International Conference on Robotics and Automation (ICRA)},
  pages={5961--5968},
  year={2025},
  organization={IEEE}
}

@article{cheng2025omnivtla,
  title={OmniVTLA: Vision-Tactile-Language-Action Models With Semantic-Aligned Tactile Sensing},
  author={Cheng, Zhengxue and Zhang, Yiqian and Tang, Anni and Wang, Keyu and Zhang, Wenkang and Li, Haoyu and Zhang, Hengdi and Song, Li},
  journal={IEEE Robotics and Automation Letters},
  year={2026},
  publisher={IEEE}
}

@inproceedings{higuera2025sparshx,
  title={Tactile beyond pixels: Multisensory touch representations for robot manipulation},
  author={Higuera, Carolina and Sharma, Akash and Fan, Taosha and Bodduluri, Chaithanya Krishna and Boots, Byron and Kaess, Michael and Lambeta, Mike and Wu, Tingfan and Liu, Zixi and Hogan, Francois Robert and others},
  booktitle={Proceedings of the Conference on Robot Learning},
  pages={105--123},
  year={2025}
}

@article{liu2025vitamin,
  title={Vitamin: Learning contact-rich tasks through robot-free visuo-tactile manipulation interface},
  author={Liu, Fangchen and Li, Chuanyu and Qin, Yihua and Xu, Jing and Abbeel, Pieter and Chen, Rui},
  journal={arXiv preprint arXiv:2504.06156},
  year={2025}
}

@article{heng2026vitacformer,
  title={Vitacformer: Learning cross-modal representation for visuo-tactile dexterous manipulation},
  author={Heng, Liang and Geng, Haoran and Zhang, Kaifeng and Abbeel, Pieter and Malik, Jitendra},
  journal={arXiv preprint arXiv:2506.15953},
  year={2025}
}

@article{krohn2026msdp,
  title={Self-Supervised Multisensory Pretraining for Contact-Rich Robot Reinforcement Learning},
  author={Krohn, Rickmer and Prasad, Vignesh and Tiboni, Gabriele and Chalvatzaki, Georgia},
  journal={IEEE Robotics and Automation Letters},
  volume={11},
  number={6},
  pages={6799--6806},
  year={2026}
}

@inproceedings{chung2023umt5,
  title={Unimax: Fairer and more effective language sampling for large-scale multilingual pretraining},
  author={Chung, Hyung Won and Constant, Noah and Garcia, Xavier and Roberts, Adam and Tay, Yi and Narang, Sharan and Firat, Orhan},
  booktitle = {International Conference on Learning Representations},
  year={2023}
}

@inproceedings{liu2022convnext,
  title={A convnet for the 2020s},
  author={Liu, Zhuang and Mao, Hanzi and Wu, Chao-Yuan and Feichtenhofer, Christoph and Darrell, Trevor and Xie, Saining},
  booktitle={IEEE/CVF conference on computer vision and pattern recognition},
  pages={11966--11976},
  year={2022},
  organization={IEEE}
}

@inproceedings{kerr2023ssvtp,
  title={Self-supervised visuo-tactile pretraining to locate and follow garment features},
  author={Kerr, Justin and Huang, Huang and Wilcox, Albert and Hoque, Ryan and Ichnowski, Jeffrey and Calandra, Roberto and Goldberg, Ken},
  booktitle={Proceedings of Robotics: Science and Systems},
  year={2023}
}

@article{bai2025qwen3,
  title={Qwen3-vl technical report},
  author={Bai, Shuai and Cai, Yuxuan and Chen, Ruizhe and Chen, Keqin and Chen, Xionghui and Cheng, Zesen and Deng, Lianghao and Ding, Wei and Gao, Chang and Ge, Chunjiang and others},
  journal={arXiv preprint arXiv:2511.21631},
  year={2025}
}

@inproceedings{cheng2025opentelevision,
  title = 	 {Open-TeleVision: Teleoperation with Immersive Active Visual Feedback},
  author =       {Cheng, Xuxin and Li, Jialong and Yang, Shiqi and Yang, Ge and Wang, Xiaolong},
  booktitle = 	 {Proceedings of the Conference on Robot Learning},
  pages = 	 {2729--2749},
  year = 	 {2025},
  volume = 	 {270}
}

@inproceedings{wang2024dexcap,
  title={Dexcap: Scalable and portable mocap data collection system for dexterous manipulation},
  author={Wang, Chen and Shi, Haochen and Wang, Weizhuo and Zhang, Ruohan and Fei-Fei, Li and Liu, C Karen},
  booktitle={Proceedings of Robotics: Science and Systems},
  year={2024}
}

@inproceedings{jiang2025dexmimicgen,
  title={Dexmimicgen: Automated data generation for bimanual dexterous manipulation via imitation learning},
  author={Jiang, Zhenyu and Xie, Yuqi and Lin, Kevin and Xu, Zhenjia and Wan, Weikang and Mandlekar, Ajay and Fan, Linxi Jim and Zhu, Yuke},
  booktitle={2025 IEEE International Conference on Robotics and Automation (ICRA)},
  pages={16923--16930},
  year={2025}
}

@inproceedings{ding2025bunnyvisionpro,
  title={Bunny-visionpro: Real-time bimanual dexterous teleoperation for imitation learning},
  author={Ding, Runyu and Qin, Yuzhe and Zhu, Jiyue and Jia, Chengzhe and Yang, Shiqi and Yang, Ruihan and Qi, Xiaojuan and Wang, Xiaolong},
  booktitle={2025 IEEE/RSJ International Conference on Intelligent Robots and Systems (IROS)},
  pages={12248--12255},
  year={2025},
  organization={IEEE}
}

@inproceedings{sun2025vtaobimanip,
  title={VTAO-BiManip: Masked visual-tactile-action pre-training with object understanding for bimanual dexterous manipulation},
  author={Sun, Zhengnan and Shi, Zhaotai and Chen, Jiayin and Liu, Qingtao and Cui, Yu and Chen, Jiming and Ye, Qi},
  booktitle={2025 IEEE/RSJ International Conference on Intelligent Robots and Systems (IROS)},
  pages={3201--3208},
  year={2025},
  organization={IEEE}
}

@article{wang2026robotacdex,
  title={RoboTacDex: A Dexterous Visual-Tactile-Action Dataset for Humanoid Manipulation},
  author={Wang, Xinyi and Li, Donghan and Chen, Zi'Ang and Yu, Chong and Xin, Chen and Ye, Peng and Sun, Yingkai and Chen, Tao},
  journal={arXiv preprint arXiv:2606.31836},
  year={2026}
}

@inproceedings{li2026deco,
  title = 	 {{DECO}: Decoupled Multimodal Diffusion Transformer for Bimanual Dexterous Manipulation with a Plugin Tactile Adapter},
  author =       {Li, Xukun and Sun, Yu and Zhang, Lei and Huang, Bo-Sheng and Peng, Yibo and Meng, Yuan and Jiang, Haojun and Xie, Shaoxuan and Yao, Guocai and Knoll, Alois and Bing, Zhenshan and Wang, Xinlong and Sun, Zhenguo},
  booktitle = 	 {Proceedings of the 43rd International Conference on Machine Learning},
  pages = 	 {70113--70133},
  year = 	 {2026},
  volume = 	 {306}
}

@article{liu2026star,
  title={STAR: Sparse Tactile Representation Learning in Vision-Tactile-Language-Action Models for Dexterous Manipulation},
  author={Liu, Xiangcheng and Wu, Tianhao and Zheng, Le and Wang, Yidong and Jiang, Bowen and Pan, Mingjie and Ren, Xinlin and Liu, Yi and Luo, Jianlan},
  journal={arXiv preprint arXiv:2609.12549},
  year={2026}
}

@inproceedings{wu2024gr1,
  title={Unleashing large-scale video generative pre-training for visual robot manipulation},
  author={Wu, Hongtao and Jing, Ya and Cheang, Chilam and Chen, Guangzeng and Xu, Jiafeng and Li, Xinghang and Liu, Minghuan and Li, Hang and Kong, Tao},
  booktitle={International Conference on Learning Representations},
  volume={2024},
  pages={10641--10662},
  year={2024}
}

@article{guo2024pad,
  title={Prediction with action: Visual policy learning via joint denoising process},
  author={Guo, Yanjiang and Hu, Yucheng and Zhang, Jianke and Wang, Yen-Jen and Chen, Xiaoyu and Lu, Chaochao and Chen, Jianyu},
  journal={Advances in Neural Information Processing Systems},
  volume={37},
  pages={112386--112410},
  year={2024}
}

@article{ye2026dreamzero,
  title={World action models are zero-shot policies},
  author={Ye, Seonghyeon and Ge, Yunhao and Zheng, Kaiyuan and Gao, Shenyuan and Yu, Sihyun and Kurian, George and Indupuru, Suneel and Tan, You Liang and Zhu, Chuning and Xiang, Jiannan and others},
  journal={arXiv preprint arXiv:2602.15922},
  year={2026}
}

@article{luo2026beingh07,
  title={Being-h0. 7: A latent world-action model from egocentric videos},
  author={Luo, Hao and Zhang, Wanpeng and Feng, Yicheng and Zheng, Sipeng and Xu, Haiweng and Xu, Chaoyi and Xi, Ziheng and Fu, Yuhui and Lu, Zongqing},
  journal={arXiv preprint arXiv:2605.00078},
  year={2026}
}

@article{fang2026lewam,
  title={Latent evolving World Action Model},
  author={Fang, Xueji and Duan, Boqiang and Wu, Hua and Wang, Jingdong and Qi, Guo-Jun},
  journal={arXiv preprint arXiv:2609.27455},
  year={2026}
}

@article{clemente2026lajepa,
  title={La-JEPA: Compact Robot Policies with Latent World Modeling},
  author={Cl{\'e}mente, Mateo and Xue, George Z and Abyaneh, Amin and Ma, Liheng and Yang, Rui Heng and Cao, Tongtong and Zhang, Zhanguang and Zhang, Yingxue},
  journal={arXiv preprint arXiv:2611.08156},
  year={2026}
}

@article{tang2026modevla,
  title={Towards Human-Like Manipulation through RL-Augmented Teleoperation and Mixture-of-Dexterous-Experts VLA},
  author={Tang, Tutian and Ji, Xingyu and Xing, Wanli and Hao, Ce and Xu, Wenqiang and Shao, Lin and Lu, Cewu and Yu, Qiaojun and Pang, Jiangmiao and Zhang, Kaifeng},
  journal={arXiv preprint arXiv:2603.08122},
  year={2026}
}

@article{motoda2026self,
  title={Self-Supervised Anchoring of Fingertip Sensing to Proprioception and Proactive Actions for Robot Imitation Learning},
  author={Motoda, Tomohiro and Murooka, Masaki and Shirai, Keisuke and Oh, Hanbit and Nakajo, Ryoichi and Miwa, Shotaro and Mykhailyshyn, Roman and Duarte, Hugo and Domae, Yukiyasu},
  journal={arXiv preprint arXiv:2609.29822},
  year={2026}
}

@article{musa2026and,
  title={What, where, and how: Probing spatiotemporal representations in video foundation models},
  author={Musa, Sharon S and Forghani, Fereshteh and Thasarathan, Harrish and Joseph, Sonia and Kowal, Matthew and Derpanis, Konstantinos G},
  journal={arXiv preprint arXiv:2609.01551},
  year={2026}
}

@inproceedings{jaegle2021perceiver,
  title={Perceiver: General perception with iterative attention},
  author={Jaegle, Andrew and Gimeno, Felix and Brock, Andy and Vinyals, Oriol and Zisserman, Andrew and Carreira, Joao},
  booktitle={International conference on machine learning},
  pages={4651--4664},
  year={2021},
  organization={PMLR}
}

@inproceedings{assran2023ijepa,
  title={Self-supervised learning from images with a joint-embedding predictive architecture},
  author={Assran, Mahmoud and Duval, Quentin and Misra, Ishan and Bojanowski, Piotr and Vincent, Pascal and Rabbat, Michael and LeCun, Yann and Ballas, Nicolas},
  booktitle={IEEE/CVF Conference on Computer Vision and Pattern Recognition (CVPR)},
  pages={15619--15629},
  year={2023},
  organization={IEEE}
}

@article{assran2025vjepa2,
  title={V-jepa 2: Self-supervised video models enable understanding, prediction and planning},
  author={Assran, Mido and Bardes, Adrien and Fan, David and Garrido, Quentin and Howes, Russell and Muckley, Matthew and Rizvi, Ammar and Roberts, Claire and Sinha, Koustuv and Zholus, Artem and others},
  journal={arXiv preprint arXiv:2506.09985},
  year={2025}
}

@inproceedings{mur2026vjepa21,
  title={V-jepa 2.1: Unlocking dense features in video self-supervised learning},
  author={Mur-Labadia, Lorenzo and Muckley, Matthew and Bar, Amir and Assran, Mido and Sinha, Koustuv and Rabbat, Mike and LeCun, Yann and Ballas, Nicolas},
  booktitle={European Conference on Computer Vision},
  pages={671--689},
  year={2026},
  organization={Springer}
}

@inproceedings{schroff2015facenet,
  title={Facenet: A unified embedding for face recognition and clustering},
  author={Schroff, Florian and Kalenichenko, Dmitry and Philbin, James},
  booktitle={Proceedings of the IEEE conference on computer vision and pattern recognition},
  pages={815--823},
  year={2015}
}

@article{lipman2022flow,
  title={Flow matching for generative modeling},
  author={Lipman, Yaron and Chen, Ricky TQ and Ben-Hamu, Heli and Nickel, Maximilian and Le, Matt},
  journal={arXiv preprint arXiv:2210.02747},
  year={2022}
}

@article{cheng2026tau,
  title={$\tau$: Learning Touch-Augmented Vision-Language-Action Models from Future Visual Supervision},
  author={Cheng, Ning and Xu, Jinan and Li, Wanlin and Chen, Yangzhi and Gao, Jing and Wang, Yiqun and Peng, Kelan and Han, Wenjuan},
  journal={arXiv preprint arXiv:2607.24485},
  year={2026}
}

@article{community2026starvla,
  title={StarVLA: A Lego-like Codebase for Vision-Language-Action Model Developing},
  author={Community, StarVLA},
  journal={arXiv preprint arXiv:2604.05014},
  year={2026}
}
}


\end{document}